\documentstyle[icassp91]{article}

\title{SPECIALIZED LANGUAGE MODELS USING DIALOGUE PREDICTIONS}

\name{\parbox{2.7in}{\centering Cosmin Popovici}
\ ~~~~~~~~ \
\parbox{3in}{\centering Paolo Baggia}}

\address{
\parbox[t]{2.7in}{\centering {\small ICI -
Institutul de Cercetari in Informatica\\
Bd. M. Averescu, 8-10\\
Bucuresti (Romania)}}
\ ~~~~~~~~ \
\parbox[t]{3in}{\centering {\small CSELT -
Centro Studi e Laboratori Telecomunicazioni\\
Via G. Reiss Romoli, 274\\
I-10148 Torino (Italy)\\
{\tt baggia@cselt.stet.it}}}}

\input{psfig}

\begin{document}

\ninept
\maketitle

\begin{abstract}

This paper analyses language modeling in spoken 
dialogue systems for accessing a database. The use of 
several language models obtained by exploiting dialogue 
predictions gives better results than the use of a single 
model for the whole dialogue interaction. For this reason 
several models have been created, each one for a specific 
system question, such as the request or the confirmation 
of a parameter.

The use of dialogue-dependent language models 
increases the performance both at the recognition and at 
the understanding level, especially on answers to system 
requests. Moreover using other methods to increase 
performances, like automatic clustering of vocabulary-
words or the use of better acoustic models during 
recognition, does not affect the improvements given by 
dia\-logue-dependent language models.

The system used in our experiments is Dialogos, the 
Italian spoken dialogue system used for accessing railway 
timetable information over the telephone. The 
experiments were carried out on a large corpus of 
dialogues collected using Dialogos.

\end{abstract}

\section{INTRODUCTION}

In a spoken dialogue system (SDS) a method to 
improve speech recognition and speech understanding is 
to use contextual knowledge as a constraint, both at the 
recognition and at the parsing level~\cite{Young89}.

Carter~\cite{Carter94} shows that clustering the sentences of the 
training corpus into subcorpora on the basis of the 
criterion of minimizing entropy, improves n-gram based 
language models. We propose that the splitting of a 
corpus acquired from a SDS should be done according to 
the dialogue point in which an utterance was given. On 
these subcorpora a set of more specific n-gram based 
language models was trained. This work extends the 
previous one described in~\cite{Gerbino95}, where first insights into
the usefulness of dialogue predictions were given on a corpus 
acquired with an earlier version of the dialogue system, 
see~\cite{Baggia94}.

Our use of dialogue prediction is similar to the static 
prediction described in~\cite{Andry92} and is related to the dialogue-
step dependent models in~\cite{Eckert96}, the difference being that we 
also measured performance at the understanding le\-vls
el.

Other methods to improve SDS performances in 
conjunction with the use of dialogue predictions were 
tested. The work developed in~\cite{Moisa95} was exploited and the 
vocab\-ulary-words (VW) were clustered automatically. 
Further improvement was obtained using acoustic models 
trained on a larger training-set of domain specific 
utterances. It's remarkable that even in those cases the 
improvements given by dialogue-dependent language 
models were not affected.

\section{THE SYSTEM USED FOR THE ACQUISITION}

Dialogos is an all-software, completely integrated, 
dialogue system which runs very close to real-time on a 
DEC Alpha, except for the telephonic interface and text-
to-speech synthesizer which are run from a PC equipped 
with a D41E Dialogic board.

The acoustical front-end performs feature extraction 
and acoustic-phonetic decoding. The recognition module 
is based on a frame-synchronous Viterbi decoding, where 
the acoustic matching is performed by a phonetic neural 
network~\cite{Gemello94}. The vocabulary of Dialogos contains 3,471 
words, clustered in 358 classes. 348 of them contain a 
single word, while the remaining 10 classes contain 
semantically important words, such as city names (2,983 
words), station names (33 words), numbers (76 words), 
months, week days, and so on. During the recognition, a 
class-based bigram language model is used. It was trained 
on 30,000 sentences. The training data of the language 
models was partially derived from a previous trial of SDS 
applied to the same domain, but for the most part (86\%) it 
was manually created.

The linguistic processor starts from the best-decoded 
sequence, and it performs a multi-step robust partial 
parsing, which is an improvement of~\cite{Rullent93}. It accepts
partial solutions on the basis of their coherence with respect to 
the parser's linguistic knowledge and generates a task-
oriented semantic caseframe.

To interpret a new utterance in the on-going 
interaction, the dialogue module (DM) takes into account 
the linguistic history and the active focus. This 
mechanism allows the DM to identify linguistic 
references, find out the correct context to apply to 
utterance interpretation, and decide if an utterance causes 
a shift or restriction of focus~\cite{Danieli93}. The result of the 
contextual interpretation is the choice of a proper 
dialogue act (DA), such as the request or the confirmation 
of a parameter, and the generation of a system answer. 
The DM makes use of pragmatic expectations about what 
the user would probably say in a certain dialogue state. 
On the basis of these contextual based expectations the 
DM can generate predictions. 

\begin{figure}[htb]
\begin{footnotesize}
\begin{tabular}{ll} \hline
$S_{1}$$>$ & Where are you leaving from and going to? \\
& {\bf $<request: departure\_city,arrival\_city>$} \\
$U_{1}$$>$ & From Turin to Milan. \\
$S_{2}$$>$ & Are you leaving from Turin for Milan? \\
& {\bf $<confirm: departure\_city,arrival\_city>$} \\
$U_{2}$$>$ & Yes tomorrow at about eight o'clock. \\
$S_{3}$$>$ & Do you want to leave tomorrow at about eight \\
    & o'clock?  \\
& {\bf $<confirm: departure\_date,arrival\_time>$} \\
$U_{3}$$>$ & Yes. \\
$S_{4}$$>$ & I have found two connections \dots Do you want \\
    & other information about these connections?	 \\
& {\bf $<confirm: connection\_information>$} \\
$U_{4}$$>$ & No thanks. \\
$S_{5}$$>$ & Thank you for the call. Good-bye. \\ \hline
\end{tabular}
\end{footnotesize}
\caption{\label{esdial} Example of a dialogue interaction.}
\end{figure}

Using Dialogos a corpus\footnote{A part of this corpus
collected from 493 naive users (1,363 
dialogues, 13,123 utterances) is reported
in~\cite{Albesano97}, where the 
evaluation results of the system are given.}
of near 2,000 dialogues for a 
total of 19,697 utterances was acquired. A dialogue 
example is shown in Figure~\ref{esdial}, where for each system 
sentence ({\bf $S_{i}>$}) the DA and the parameters are given. This 
information can also be used for predicting a more 
specific language model which better represents the 
syntactic, semantic, and contextual constraints of the 
future user's answer.

\section{PREDICTIONS}

The concept of prediction constitutes the guessing of a 
future action and it is commonly used to obtain 
constraints in a certain point of a dialogue. In an 
information inquiry system the knowledge to estimate the 
subset of user's DA already exists. In the VERMOBIL 
system~\cite{Reithinger95}, for instance, a special module estimates
the set of DAs in the next user utterance and a stochastic 
recovery is done when the prediction fails. In our system 
a certain point in a dialogue is identified by the question 
that the user is replying to, i.e. the DA of the system 
generated sentence, which is called in the following 
dialogue prediction (DP).

At the recognition level, we make use of the 
information that the DM can provide, by creating specific 
LMs for each DP. The most specific LM is obtained from 
a training-set which only contains replies given in a 
certain DP. However, some questions very rarely appear 
and for them the information contained in the training DB 
is not enough to obtain a robust LM.

\subsection{Question classification}

\begin{figure}[htb]
\psfig{figure=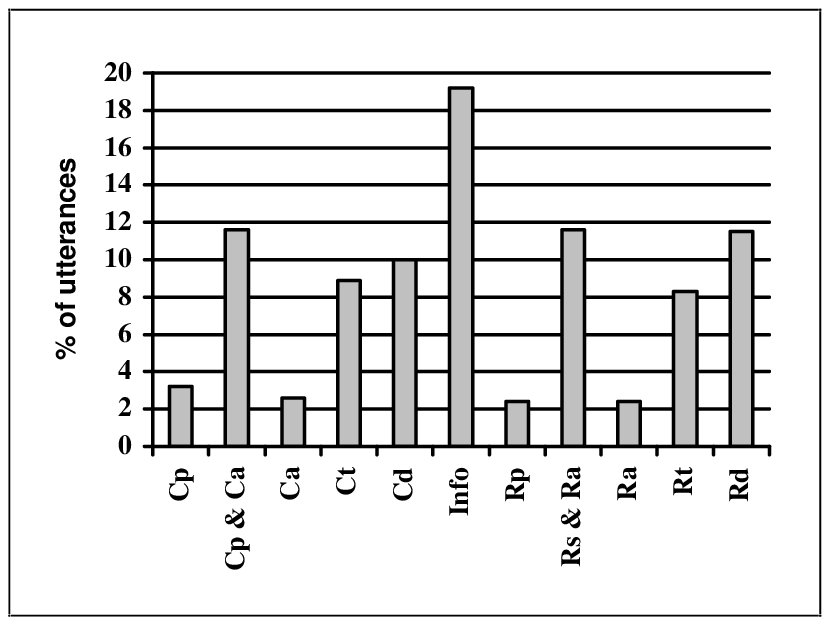,width=7.9cm}
\caption{\label{freq}Relative frequencies of the classes.}
\end{figure}

The system questions were classified in a natural way. 
At first they were divided into groups according to the 
type of DA: request for ({\bf $R_{i}$}) and confirmation
of ({\bf $C_{i}$}) a 
parameter {\bf $i$}, and listing of train information
({\bf $Info$}). Then 
these groups were separated into DAs involving one or 
more parameters, and, finally, a distinction was made 
between the different parameters dealt with by the 
questions, such as departure city ({\bf $p$}), arrival city ({\bf $a$}), 
departure time ({\bf $t$}), and departure date ({\bf $d$}). For example, 
{\bf $C_{p}$} is the confirmation of the departure city,
{\bf $R_{t}$} the 
request of the departure time, and {\bf $R_{p}$}~\&~{\bf $R_{a}$} the
request of both the departure and the arrival cities through a single 
sentence. In Figure~\ref{freq} the various classes are shown 
together with the frequencies of occurrence in the 
acquired corpus.

Bearing in mind these distinctions, a specific training
set for each class was obtained. The utterances of a 
specific training-set include all the instances of different 
user's answers in that point of the dialogue, for instance in 
the {\bf $C_{p}$} training-set there are both positive and negative 
confirmations.

\subsection{Creation of the models}

After obtaining the training-sets for each specific 
class, different models were created with the same 
algorithm used for a single context-independent model. 
All the results presented in this paper were obtained using 
both a bigram model during the acoustic decoding and a 
trigram one for the rescoring of the 25 n-best sequences.

\section{EXPERIMENTAL RESULTS}

We carried out two sets of experiments using either a 
single model for all utterances or a set of specific models 
that takes into account the predictions described before. 
Both the context-independent and the specialized models 
were trained on the same material, 15,575 user utterances, 
and tested on 2,040 ones. The two sets were disjunctive. 
Performance is measured at both recognition and 
understanding levels. Recognition performance is 
measured in terms of sentence accuracy (SA) and word 
accuracy (WA), and understanding one in terms of 
sentence understanding (SU\footnote{SU is obtained comparing
for each sentence the caseframe 
generated by the parser with a manually corrected one. The CA 
takes into account substitution, insertion, and deletion of 
concepts, i.e. attibute-value pairs in the caseframe. The CA 
formula is similar to the WA one, see~\cite{Boros96}.})
and concept accuracy (CA).

\subsection{Single context-independent models}

Table~\ref{tab1} shows the comparison of the performance of 
the LM used during the acquisition (baseline) and a single 
dialogue-independent LM obtained with the whole 
training-set (ALL\_INT). The baseline model was mainly 
trained on manually created data, which some of them are 
unusual in a dialog interaction, and so this model shows a 
poor level of specificity. The ALL\_INT model, on the 
other hand, is far more specific, as it only includes 
utterances occurred through the user dialogues, and so it 
reflects the distribution of the utterances in a real setting. 
Both at the recognition and the understanding levels the 
ALL\_INT model gives a better performance. 

\begin{table}[htb]
\begin{center}
\begin{tabular}{|l|c|c|c|c|} \hline
            &   SA      & WA    &   SU &  CA \\ \hline
baseline    &   69.4    & 68.8  & 76.1 &  66.4 \\ \hline
ALL\_INT     &   70.9    & 71.1  & 77.6 &  68.5 \\ \hline
ALL\_PRED    &   71.2    & 73.1  & 79.4 &  72.2 \\ \hline
FINAL       &   71.5    & 73.4  & 79.8 &  72.5 \\ \hline
\end{tabular}
\end{center}
\caption{\label{tab1}Results of single models and models with DP.}
\end{table}

\subsection{Language models with dialogue predictions}

A set of two models with DP were tested. The first 
one, ALL\_PRED, was created as described in Section 
3.2. Another one, FINAL, takes for each class the best 
between the single model (ALL\_INT) and  the model 
with DP (ALL\_PRED), according to the SU metric. For 
classes containing a few utterances the ALL\_INT model 
was preferable, for instance, in the class ``confirmation of 
departure city'' (Cp), so in this case it was selected.
The results for the models with DP are also given in 
Table~\ref{tab1}. They show that the use of DP almost double the 
improvement obtained with the ALL\_INT model alone. 
The error rate reduction between ALL\_INT and FINAL is 
near 10\% for WA and SU, and over 20\% for CA. These 
improvements are encouraging because they compare 
favorably with the ones reported in~\cite{Eckert96}.

The improvements became clearer if we separate the 
test utterances into requests for and confirmations of a 
parameter, as shown in Table~\ref{tab2}. Through the use of DP 
(the FINAL model) a general improvement for the 
request utterances of 2-4\% was achieved. This was 
slightly reduced for the confirmations, because about 
70\% of them are utterances of only one word ("Yes", 
"No", "Okay", and so on), which are always correctly 
recognized.

\begin{table}[htb]
\begin{center}
\begin{tabular}{|l|l|c|c|c|c|} \hline
            &         &   SA      & WA    &   SU &  CA   \\ \hline
request     & ALL\_INT &   60.8    & 74.6  & 67.4 &  60.6 \\ \hline
request     & FINAL   &   62.8    & 78.9  & 71.3 &  66.3 \\ \hline
confirm     & ALL\_INT &   77.3    & 71.9  & 84.6 &  76.5 \\ \hline
confirm     & FINAL   &   76.9    & 71.3  & 85.4 &  78.1 \\ \hline
\end{tabular}
\end{center}
\caption{\label{tab2}Results for requests and confirmations.}
\end{table}

\section{PREDICTIONS VS. OTHER IMPROVEMENTS}

It is interesting to test if the increment of performance 
brought by the use of DP is affected by the use of other 
methods. Two methods were tested, such as: the 
automatic clustering of vocabulary words (ACVW) and 
the use of acoustic models trained on a larger set of 
domain specific utterances.

\subsection{Language models with automatic clustering of 
vocabulary words}

Word clustering is commonly used to reduce number 
of parameters of a LM. This could increase the statistical 
robustness and reduce the size of the model itself.
At first, most of the classes (348 from 358) had one 
single word, and these classes were clustered again in 
automatic way using Maximum likelihood
method\footnote{In~\cite{Moisa95}
several clustering methods were compared through the 
perplexity values and they gave similar results. In this work the 
choice of the best automatic clustering method was made 
experimentally.}, as 
described in~\cite{Moisa95}. The final number of classes was 120.
Two models FINAL-clust, and ALL\_INT-clust were 
trained on the same database as FINAL, and ALL\_INT 
described above, but the word classification was changed 
from 358 to 120 classes.

\subsection{Use of more specific acoustic models}

All experimental results till now, have used an 
acoustic model (M1) trained on a set of two DBs. The 
first is a domain independent one, which contains 
phonetically balanced data produced by 1,136 speakers, 
4,875 utterances (with an average length of 6 words) and 
3,653 isolated words. The second one is domain 
dependent, and it includes 3,580 utterances (with an 
average length of 2 words) from 270 speakers. It came 
from an older SDS acquisition.
A new acoustic model (M2) was created by adding 
13,929 utterances (with an average length of 2 words), 
from the corpus described in Section 2, to the domain 
dependent DB part of M1.

\subsection{Final comparison}

Table~\ref{tab3} shows WA and SU results for the LMs with 
autoclassification using both M1 and M2 acoustic models. 
Autoclassfication only (M1 columns) improved both the 
single model and the DP one, compared to the results in 
Table~\ref{tab1}, and, as expected, the M2 acoustic models 
furtherly increment the recognition and understanding 
results. In any case these improvements does not alter the 
advantage obtained by the use of DP.

\begin{table}[htb]
\begin{center}
\begin{tabular}{|l|c|c|c|c|} \hline
 & \multicolumn{2}{|c|}{WA} & \multicolumn{2}{|c|}{SU} \\ \hline
                  &   M1      & M2    & M1   &  M2   \\ \hline
ALL\_INT-clust    &   71.9    & 73.8  & 79.0 &  81.4 \\ \hline
FINAL-clust       &   73.4    & 75.6  & 80.8 &  83.5 \\ \hline
\end{tabular}
\end{center}
\caption{\label{tab3}Comparison between models with ACVW.}
\end{table}

\begin{figure}[htb]
\psfig{figure=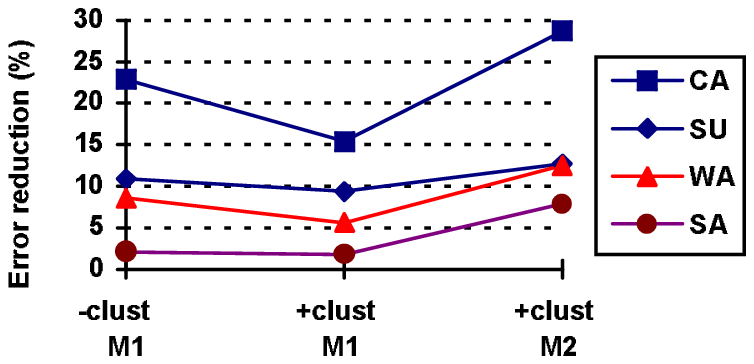}
\caption{\label{errorred}Error reduction among all the 
experimental settings.}
\end{figure}

The diagram in Figure~\ref{errorred} represents the error rate 
reduction values between ALL\_INT and FINAL LMs, for 
three different experimental settings, which are: without 
ACVW using M1 (-clust/M1); with ACVW still using M1 
(+clust/M1); and with ACVW but using M2 (+clust/M2).
The diagram shows clearly that in each case the LMs 
which use DP give better recognition and understanding 
results (all the error rate reduction values are positive). 
It's also remarkable that the use of DP, in conjunction 
with other methods, could even increase the 
improvement. All the values of +clust/M2 are greater 
then the -clust/M1 ones, so for SU it goes from 10.9\% 
(for -clust/M1) to 12.7\% (for +clust/M2), and from 22.9\%
to 28.7\% for CA. However in +clust/M1 the error 
reduction is the smallest, because the ACVW improve 
above all the single model (ALL\_INT-clust).

\section{CONCLUSIONS}

It has been shown that more specific models (created 
exclusively with replies given at a certain point of the 
dialogue) improve globally the performance of SDS. On 
the other hand, in some cases the specific models are not 
robust enough (i.e. very rare, but appropriate utterances). 
The trade-off between specificity and robustness should 
be better studied in future.
The improvement of the performance for requests 
suggests a proportional general improvement of the whole 
system, because it implies a higher number of positive 
replies to the following confirmation and the reduction of 
the number of turns in the dialogue for some unnecessary 
recovery. Moreover the use of DP is useful in conjunction 
with other methods, such as the autoclassification of 
vocabulary words and the use of more specific acoustic 
models. These kind of dialogue-dependent LMs have 
been already integrated into Dialogos system.

\begin{small}

\end{small}

\end{document}